# Utilizing Speech Emotion Recognition and Recommender Systems for Negative Emotion Handling in Therapy Chatbots

Farideh Majidi  
Computer Engineering Department  
Islamic Azad University, South Tehran Branch  
Tehran, Iran  
St_f.majidi@azad.ac.ir

Marzieh Bahrami  
Computer Engineering Department  
Islamic Azad University, South Tehran Branch  
Tehran, Iran

*Abstract*—Emotional well-being significantly influences mental health and overall quality of life. As therapy chatbots become increasingly prevalent, their ability to comprehend and respond empathetically to users' emotions remains limited. This paper addresses this limitation by proposing an approach to enhance therapy chatbots with auditory perception, enabling them to understand users' feelings and provide human-like empathy. The proposed method incorporates speech emotion recognition (SER) techniques using Convolutional Neural Network (CNN) models and the ShEMO dataset to accurately detect and classify negative emotions, including anger, fear, and sadness. The SER model achieves a validation accuracy of 88%, demonstrating its effectiveness in recognizing emotional states from speech signals. Furthermore, a recommender system is developed, leveraging the SER model's output to generate personalized recommendations for managing negative emotions, for which a new bilingual dataset was generated as well since there is no such dataset available for this task. The recommender model achieves an accuracy of 98% by employing a combination of global vectors for word representation (GloVe) and LSTM models. To provide a more immersive and empathetic user experience, a text-to-speech model called Glow-TTS is integrated, enabling the therapy chatbot to audibly communicate the generated recommendations to users in both English and Persian. The proposed approach offers promising potential to enhance therapy chatbots by providing them with the ability to recognize and respond to users' emotions, ultimately improving the delivery of mental health support for both English and Persian-speaking users.

*Keywords—speech emotion recognition, recommender system, therapy chatbot, deep learning*

I. INTRODUCTION

Emotional well-being is a fundamental component of mental health and plays a pivotal role in shaping individuals' overall quality of life. In an era where digital technologies are increasingly interwoven into the fabric of our daily lives, the advent of therapy chatbots offers a novel avenue for addressing emotional well-being and mental health support. These automated conversational agents hold the promise of delivering empathetic and effective mental health care to a broader spectrum of individuals. However, a significant challenge that therapy chatbots currently face is their limited ability to comprehend and respond empathetically to the complex spectrum of human emotions that appear in the voice, rather than the written sentences chatbots usually receive.

In recognition of this critical issue, this paper embarks on a journey to augment the capabilities of therapy chatbots by endowing them with auditory perception, enabling them to better understand and respond to users' emotions with human-like empathy which can't be achieved from the emotions users put into words. It can also be antipathetic if the feeling in the voice of users is not perceived throughout the therapy session, which is totally of remarkable importance to real-world therapists. Our approach hinges on the integration of state-of-the-art techniques, namely Speech Emotion Recognition and recommender systems, into the framework of therapy chatbots. Through this integration, we aim to bridge the gap between technology and the nuanced realm of human emotions, with a particular focus on serving both English and Persian-speaking users.

The cornerstone of our approach lies in the utilization of Convolutional Neural Network models in tandem with the ShEMO dataset, enabling our therapy chatbots to accurately detect and classify a range of negative emotions, including anger, fear, and sadness. The output of this model represents the emotions in the user's voice which will be fed to the recommender system mode. We introduce a novel recommender system that leverages the output of the SER model, which is the negative feeling in the user's voice. This recommender system harnesses a bilingual dataset specifically created for the task, a pioneering endeavor given the absence of such datasets for this purpose. The recommender system, employing a combination of Global Vectors for Word Representation and LSTM models, achieves an astounding result in choosing recommendations for the specific feeling that the user is experiencing. In our relentless pursuit of a more immersive and empathetic user experience, we integrate the cutting-edge text-to-speech model, Glow-TTS. This addition empowers the therapy chatbot to audibly communicate the generated recommendations to users in both English and Persian, enhancing the therapeutic interaction. In fact, a text-to-speech paradigm can open up new possibilities for chatbots to be more audio-based than text-based.

Our proposed approach represents a promising milestone in the evolution of therapy chatbots. By equipping these intelligent companions with the capacity to recognize and respond to users' emotions, we aspire to elevate the standard of mental health support delivery, fostering a more empathetic and effective therapeutic alliance for both English and Persian-speaking users.

The rest of the paper is structured as follows. Section II introduces related works. Section III presents the proposed model and methodology. Section IV shows the experimental results and provides a discussion. Finally, the paper is concluded in Section V.

## II. RELATED WORKS

A study [1] introduces an innovative emotion recognition system that harnesses the power of deep learning, drawing insights from emotional Big Data encompassing both speech and video modalities. The system's pipeline begins with the preprocessing of speech signals in the frequency domain, leading to the extraction of Mel-spectrograms, which are treated as image-like data. These Mel-spectrograms are then input into a CNN model. Simultaneously, for video signals, select frames are sampled from a video segment and likewise fed into a CNN. The outputs from both CNNs are skillfully integrated using a tandem of consecutive extreme learning machines (ELMs). Subsequently, the fused output undergoes final classification using a support vector machine to ascertain emotional states.

In Another interesting work [2] the authors address the global significance of depression as a complex mental health concern. They highlight the challenges associated with recognizing and managing depression due to its intricate nature and long-lasting effects. Leveraging the opportunities presented by e-commerce and intelligent recommender systems, the paper introduces the emHealth system—a novel intelligent health recommendation system with a focus on predicting and managing depression. The paper outlines the system's architecture, involving personalized mobile apps for emotional data collection and the identification of key depression indicators. Using decision tree and support vector machine algorithms, the paper develops depression prediction models and offers personalized recommendations to guide users toward emotional well-being.

In [3], Zygadło et al. focused on sentiment and emotion recognition for English and Polish texts, aiming to develop a therapeutic chatbot. They created a bilingual corpus called CORTEX, labeled with sentiment polarity and emotion classes, and employed classifiers like Naïve Bayes, Support Vector Machines, fastText, and BERT. Their best-performing models, based on BERT, achieved an accuracy of over 90% for sentiment classification and almost 80% for emotion classification. While their study focused on text-based emotion recognition, it provides insights into classification approaches for emotional content.

In [4], a noteworthy research, Lee et al. proposed a communication system with speech emotion recognition for companion robots. They used sound data enhancement methods and transformed speech into spectrograms using MFCC. Their CNN model, based on Google, achieved a recognition accuracy of 79.81% for five emotions (peace, happiness, sadness, anger, and fear). They also incorporated semantic analysis using the Seq2Seq framework of recurrent neural networks for chat conversations. Their work highlights the integration of speech emotion recognition into companion robot systems.

In [5], Yazdani et al. focused on emotion recognition in Persian speech using deep neural networks. They employed various deep learning techniques on the Sharif Emotional Speech Database (ShEMO), achieving an unweighted accuracy of 65.20% and a weighted accuracy of 78.29%. Their study contributes to the understanding of important factors in Farsi language emotion recognition.

In another paper [6], Qayyum et al. developed a CNN-based speech emotion recognition system. Their model achieved a convincing accuracy of 83.61% on a specific dataset, surpassing other similar tasks by a significant margin. Their work showcases the potential of CNNs in speech emotion recognition and their relevance in developing conversational and social robots.

In the area of recommender systems and emotion recognition fusion, the recommender system introduced in [7] which is a music recommendation system based on emotions reached 97.82% accuracy. Additionally, another emotion recognition and recommender system for music recommendation proposed in [8] achieved an accuracy of 86.98% for emotion recognition and 87.2% for the recommendation system they developed.

## III. METHODOLOGY

The objective of our research is to recognize the emotion in the user's voice and inform the chatbot of how the user is feeling at that precise instant so that it won't continue the discussion without taking the user's emotional state into account. In fact, understanding the user's potential experiences makes the discourse more human and empathic, which improves therapeutic sessions. Additionally, the chatbot can use the emotion it has detected as a new input to carefully choose a phrase via the recommender system to comfort the distressed user with calming suggestions regarding the situation. On top of that, the chatbot can verbally transmit not only the suggestions but the entire dialogue by using a text-to-speech model. As a result, both the user and chatbot will be speaking instead of writing, which will make the chatbot better able to understand the user's emotions and make the user feel more like they are speaking to a real therapist.

First and foremost, we focus on the crucial task of speech emotion recognition, which plays a pivotal role in enhancing the empathetic capabilities of therapy chatbots by understanding the emotions only conveyed by the user's voice. To achieve this goal, we utilized the Sharif Emotional Speech Database (ShEMO) [9], notable for its extensive collection of Persian speech recordings (3000 records) encompassing various emotional expressions. This dataset features diverse voices, including both male and female speakers, expressing emotions such as sadness, anger, happiness, surprise, fear, and neutrality. However, our study specifically concentrates on negative emotions, namely anger, fear, and sadness, as they hold particular relevance for gauging patients' emotional well-being during therapy sessions, allowing the chatbot to adapt and respond effectively. To address the SER task, we employ a convolutional neural network architecture, as illustrated in Figure 1. This choice is grounded in the network's capacity to capture intricate patterns and representations from input speech signals. By harnessing deep learning techniques, our methodology aims to extract meaningful features and discern emotional cues embedded within voice recordings, thereby enabling the therapy chatbot to accurately identify and understand all of the six emotions expressed by users. Importantly, our approach ensures that the SER model can recognize emotions expressed in both Persian and English languages, given the general consistency in emotional expression across different languages.

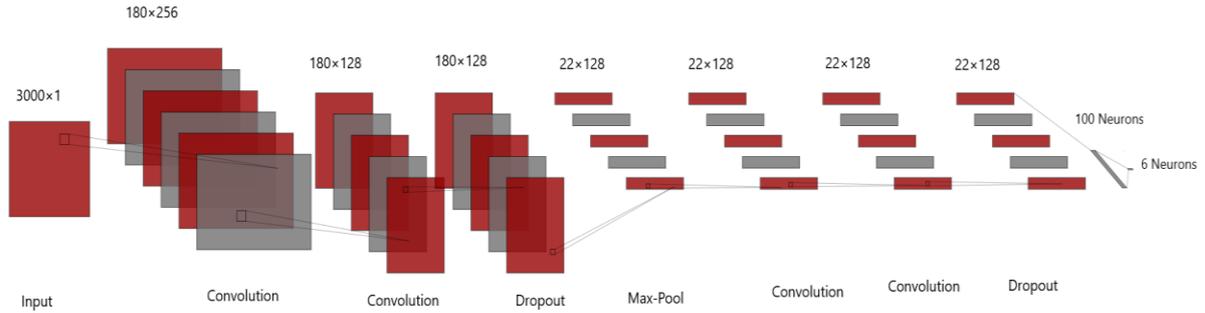

Fig. 1. CNN architecture for SER. After three convolution layers, the architecture generates six outputs, representing the emotions (happiness, surprise, neutrality, sadness, anger, and fear)

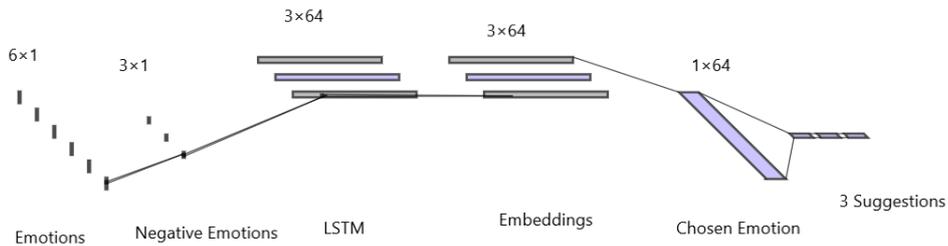

Fig. 2. LSTM architecture of recommender system. After choosing only negative emotions (anger, fear, and sadness) from all the emotions, the LSTM structure makes the decision to choose three suggestions for that particular emotion from our dataset

This versatility guarantees that the chatbot can effectively interpret and respond to emotional cues regardless of the user's language, ensuring widespread applicability and inclusivity. By leveraging the ShEMO dataset and adopting a deep neural network architecture with convolutional layers, our methodology advances the empathetic capabilities of therapy chatbots, enabling them to accurately identify and respond to emotions expressed in users' voices, ultimately enhancing the therapeutic experience. This comprehensive approach has the potential to empower therapy chatbots by recognizing and addressing users' emotional states effectively.

In the subsequent phase of our research, we focus on developing a robust recommender system capable of offering personalized recommendations based on the mood detected by the SER model. Given the absence of an existing dataset for this purpose, we create a comprehensive dataset comprising 2,500 pairs of sentences in both English and Persian languages. Each pair contains suggestions for activities or actions to undertake when experiencing specific emotions, enabling the recommender system to cater to a diverse range of users. To construct the recommender system, we employ the global vectors for word representation (GloVe) technique [10], well-known for its ability to represent words as continuous vector embeddings that capture semantic relationships and meaning between words. GloVe enhances the system's understanding of semantic similarities and connections between different suggestions within the dataset, facilitating the provision of meaningful and relevant recommendations based on the detected mood. To capture dependencies and contextual information present in our dataset, we leverage a Long Short-Term Memory (LSTM) model, a type of recurrent neural network (RNN) designed for sequential data modeling. The LSTM model enables us to capture the sequential nature of sentences, facilitating the generation of coherent and contextually appropriate suggestions. Our recommender system's architecture, depicted in Figure 2, comprises multiple layers, beginning with an input layer receiving the mood output from the SER model as the initial input for recommendation generation. Subsequently, the input passes through a GloVe embedding layer, which transforms words into vector representations, followed by the LSTM layer, which captures the contextual information and dependencies within sentences. The output layer selects three recommendations based on the processed information. Our comprehensive architecture, combining GloVe word embeddings with LSTM modeling, excels in providing accurate and contextually appropriate suggestions. By integrating mood information from the SER model, the system tailors recommendations for users based on their emotional state. Incorporating both English and Persian languages in our dataset ensures inclusivity and enables the recommender system to cater to a broader user base. This methodology lays the foundation for therapy chatbots to deliver impactful and tailored recommendations, augmenting their role as supportive companies in enhancing emotional well-being.

To enhance the practicality and user experience of our therapy chatbot framework, we seamlessly integrate a text-to-speech (TTS) model known as Glow-TTS [11]. This integration enables the conversion of recommendations generated by the recommender system into spoken audio, enhancing the interaction's naturalness and engagement. Glow-TTS utilizes advanced techniques such as neural vocoders and speech synthesis to produce high-quality and intelligible speech. The model has been trained on both Persian and English datasets, enabling it to read sentences

Fluency in both languages and accommodating users who prefer either language.

In conclusion, recommendations can be delivered audibly in either English or Persian based on the user's preference. When the SER model detects a negative emotion in the user's voice, the chatbot responds with a verbal notification, expressing empathy and acknowledging the user's emotional state. Subsequently, the related suggestion tailored to address the specific negative emotion is conveyed audibly to the user using the Glow-TTS model. This auditory delivery of recommendations aims to create a comforting and supportive experience, engaging users on a multisensory level that can enhance the effectiveness and impact of the therapeutic interaction. By incorporating the Glow-TTS model into our therapy chatbot framework, we improve functionality and user experience, providing audible recommendations that add depth and engagement to the interaction, and fostering a more empathetic and personalized therapeutic experience. This last step is integrated into our model for yet another reason which is emphasizing that it's a necessity that therapy chatbots and users talk rather than write since the important emotions in the user's voice shouldn't be ignored and the audible response from the chatbot can make the conversation more human and comfortable for the distressed patient.

## IV. RESULTS AND DISCUSSION

The results provided in Table I indicate that the model architecture for speech emotion identification and recommender system performed magnificently. The results convincingly demonstrate that our suggested model is capable of successfully identifying the emotion in the user's voice and can provide appropriate and pertinent advice to comfort the user while delivering an audio file reading the suggestions aloud.

In comparison to the mentioned related works, our proposed method incorporates SER techniques using the ShEMO dataset to accurately detect and classify negative emotions in speech signals. Our SER model achieves a validation accuracy of 88%, demonstrating its effectiveness while surpassing the other models like [5] on the same dataset, or [2] and [4] in which the authors also used a CNN-based model. Our proposed speech emotion recognition model also beats the Bert model introduced in [1] since it achieves a higher accuracy in emotion classification. We then developed a recommender system that utilizes the SER model's output to generate personalized recommendations for managing negative emotions. Our recommender model achieves an accuracy of 98% by leveraging global vectors for word representation and LSTM models, which is superior to the recommender system introduced in [7], a music recommendation system based on emotions reaching 97.82% accuracy. Additionally, another emotion recognition and recommender system for music recommendation in [8] achieved an accuracy of 86.98% for emotion recognition and 87.2% for the recommendation system they used which is also lower than our developed models' accuracies in both tasks.

## V. CONCLUSION

In conclusion, this research presents a comprehensive and innovative framework for enhancing therapy chatbots

TABLE I. SUGGESTIONS BY OUR RECOMMENDER SYSTEM

| Emotion | Suggestions (Recommender System's outputs) | | |
|---|---|---|---|
| Anger | Take a deep breath | Talk to a supportive friend | Listen to calming music |
| Sadness | Take a walk | Make a list of things you're grateful for | Think of the people you love |
| Fear | Think about the times you have overcome fear | Channel fear into positive actions | In case things get out of hand, make an emergency call |

Through speech emotion recognition, recommendation provision, and auditory perception. By leveraging SER technique on the ShEMO dataset, our proposed approach achieves a remarkable validation accuracy of 88% in detecting and classifying negative emotions, such as anger, fear, and sadness. The output of this model is the input of a recommender system that gives suggestions to the user in order to comfort them and make the conversation more human and empathic. The recommender system built upon the SER model provides highly accurate personalized recommendations from the dataset we generated which contains comforting sentences in both Persian and English to make the model multilingual and usable for a wider range of users. By achieving an impressive accuracy of 98% by utilizing GloVe and LSTM models, this model is the best fit for the task while providing meaningful and helpful suggestions. Furthermore, the integration of Glow-TTS adds an auditory dimension to the therapy chatbot, enabling the delivery of recommendations through spoken audio and enhancing the user experience. This also helps add another human aspect to the chatbots which can be an inspiration to make audible chatbots that can talk more like a real-world therapist. This research significantly contributes to the advancement of therapy chatbots by empowering them with the capability to understand and respond to users' emotions effectively rather than talking to them regardless of how they're feeling in the session. The accurate recognition of negative emotions and the provision of tailored recommendations offer valuable support and guidance to individuals seeking emotional assistance and can be a means to a more human-like therapy session. By bridging the gap between technology and emotional well-being, therapy chatbots equipped with auditory perception hold immense potential to improve the accessibility and effectiveness of mental health support.

Future research directions can further enhance the proposed framework. Expanding the model to incorporate positive emotions would provide a more comprehensive emotional support system, catering to a wider range of emotional states. Additionally, validating the models on larger and more diverse datasets would ensure the robustness and generalizability of the developed system, making it more adaptable to different contexts and user populations. Furthermore, generating new sentences with an expanded dataset can lead to the generation of more diverse and contextually relevant suggestions, enhancing the quality of recommendations provided by the therapy chatbot. Furthermore, conducting user feedback and user experience evaluations would offer valuable insights into the system's performance and usability, enabling iterative improvements and refining the overall user experience. This user-centric approach ensures that therapy chatbots are continuously refined and optimized to meet the evolving needs and expectations of users.

REFERENCES

[1] M. S. Hossain and G. Muhammad, "Emotion recognition using deep learning approach from audio–visual emotional big data," Information Fusion, vol. 49, pp. 69–78, Sep. 2019.

[2] S. Yang, P. Zhou, K. Duan, Md. S. Hossain, and M. F. Alhamid, "EMHealth: Towards Emotion Health through Depression Prediction and Intelligent Health Recommender System," *Mobile Networks and Applications*, vol. 23, no. 2, pp. 216–226, Sep. 2017.

[3] A. Zygadło, M. Kozlowski, and A. Janicki, "Text-Based emotion recognition in English and Polish for therapeutic chatbot," *Applied Sciences*, vol. 11, no. 21, p. 10146, Oct. 2021.

[4] M.-C. Lee, S.-Y. Chiang, S.-C. Yeh, and T.-F. Wen, "Study on emotion recognition and companion Chatbot using deep neural network," *Multimedia Tools and Applications*, vol. 79, no. 27–28, pp. 19629–19657, Mar. 2020.

[5] Yazdani, Ali, Hossein Simchi, and Yasser Shekofteh. "Emotion recognition in persian speech using deep neural networks." *2021 11th International Conference on Computer Engineering and Knowledge (ICCKE)*. IEEE, 2021.

[6] Qayyum, Alif Bin Abdul, Asiful Arefeen, and Celia Shahnaz. "Convolutional neural network (CNN) based speech-emotion recognition." *2019 IEEE international conference on signal processing, information, communication & systems (SPICSCON)*. IEEE, 2019.

[7] Nambiar, Kirti R., and Suja Palaniswamy. "Speech Emotion Based Music Recommendation." 2022 3rd International Conference for Emerging Technology (INCET). IEEE, 2022.

[8] T.-Y. Kim, H. Ko, S.-H. Kim, and H.-D. Kim, "Modeling of Recommendation System Based on Emotional Information and Collaborative Filtering," *Sensors*, vol. 21, no. 6, p. 1997, Mar. 2021.

[9] O. M. Nezami, P. J. Lou, and M. Karami, "ShEMO: a large-scale validated database for Persian speech emotion detection," *Language Resources and Evaluation*, vol. 53, no. 1, pp. 1–16, Oct. 2018.

[10] Pennington, Jeffrey, Richard Socher, and Christopher D. Manning."Glove: Global vectors for word representation." Proceedings of the2014 conference on empirical methods in natural language processing (EMNLP). 2014.

[11] J. Kim, S.-W. Kim, J. Kong, and S. Yoon, "Glow-TTS: a generative flow for Text-to-Speech via monotonic alignment search," *arXiv (Cornell University)*, May 2020.

[12] B. Liu and S. S. Sundar, "Should Machines Express Sympathy and Empathy? Experiments with a Health Advice Chatbot," *Cyberpsychology, Behavior, and Social Networking*, vol. 21, no. 10, pp. 625–636, Oct. 2018.

[13] V. Dhanasekar, Y. Preethi, S. Vishali, P. J. I. R, and B. P. M, "A Chatbot to promote Students Mental Health through Emotion Recognition," *2021 Third International Conference on Inventive Research in Computing Applications (ICIRCA)*, Sep. 2021.

[14] Z. Jian-Feng, X. Mao, and L. Chen, "Speech emotion recognition using deep 1D & 2D CNN LSTM networks," *Biomedical Signal Processing and Control*, vol. 47, pp. 312–323, Jan. 2019.

[15] A. Bastanfard and A. Abbasian, "Speech emotion recognition in Persian based on stacked autoencoder by comparing local and global features," *Multimedia Tools and Applications*, vol. 82, no. 23, pp. 36413–36430, Mar. 2023.